\documentclass[runningheads]{llncs}
\usepackage[T1]{fontenc}
\usepackage{graphicx}
\usepackage{amsmath}
\begin{document}
\title{Spotting the Artificial: Dialogic Humor Improvisation in Humans and LLMs}

%
%
\author{Lizeth Palacios Patiño \inst{1} Ximena Gutierrez-Vasques \inst{2} Daniel Rodriguez Vergara\inst{1}}
\institute{
  Posgrado de lingüística, UNAM, Mexico \\
  \and
  CEIICH, UNAM, Mexico \\
  \email{lizethpp.30@gmail.com xim@unam.mx danielrv@unam.mx }
}

\authorrunning{Palacios et al.}
%

\maketitle              
\begin{abstract}
\setcounter{footnote}{0}
We propose a methodology for investigating the differences between natural and synthetic language in a challenging setting: humorous improvisational dialogue. Our experimental setting compares human-produced comedy improvisations with the dialogues generated by LLMs in a multi-agent environment that we designed. The analysis combines automated linguistic measures with manual annotation grounded in Systemic Functional Linguistics (SFL). 

The results show that, although LLMs can create coherent and creative scenes, they often fail to reproduce key characteristics of spontaneous spoken interaction. In addition, LLM-generated interactions display a strong tendency toward politeness and positive evaluations, contrasting with the broader range of attitudes found in human humorous improvisation. Our evidence highlights both the strengths and limitations of current LLMs in modeling dialogic humor and demonstrate the value of integrating linguistic theory with NLP methods for the analysis of machine-generated language.\footnote{Code and data can be found at: \url{https://github.com/ximenina/DialogicHumor}}

\keywords{Humor, LLMs, functional linguistics, dialogue}

\end{abstract}

\section{Introduction}

The artificial modeling of natural language has been a central challenge in computational linguistics, as well as in AI, since the field very first origins. Approaches have evolved from formal and symbolic methods to statistical techniques, culminating in the recent emergence of large language models (LLMs). These models have brought us closer than ever to the possibility of computers that can effectively process, analyze, and generate natural language. This development is significant not only because of the wide range of applications enabled by generative artificial intelligence, but also from a linguistic perspective: What mechanisms allow a computer to model language so successfully? And what are the defining characteristics of this "synthetic" manifestation of human language?

To assess the extent to which these manifestations of language (synthetic and natural) are similar, and to identify the features that distinguish them, one must examine a truly challenging setting in which language generation requires the integration of multiple linguistic dimensions. This setting demands not only the mastery of grammatical structures, but also the ability to handle the semantic, pragmatic, and dialogic aspects of language use.

To this end, in this work we analyze language in a setting of \textbf{spontaneous humorous improvisation}. This context requires, on the one hand, dialogic interaction among multiple participants. However, these interactions are not merely conversational: they must be spontaneous, as the dialogue is improvised in real time rather than premeditated, requiring participants to continuously build upon and respond to one another's contributions. On the other hand, the generation of humor constitutes a particularly challenging linguistic task, as it relies on the successful handling of semantic, pragmatic, and cultural cues \cite{attardo2008semantics}. As such, humorous improvisation provides an ideal setting for examining the complex mechanisms involved in natural language generation and interaction. 

\begin{figure}[t]
    \centering
    \includegraphics[scale=0.14]{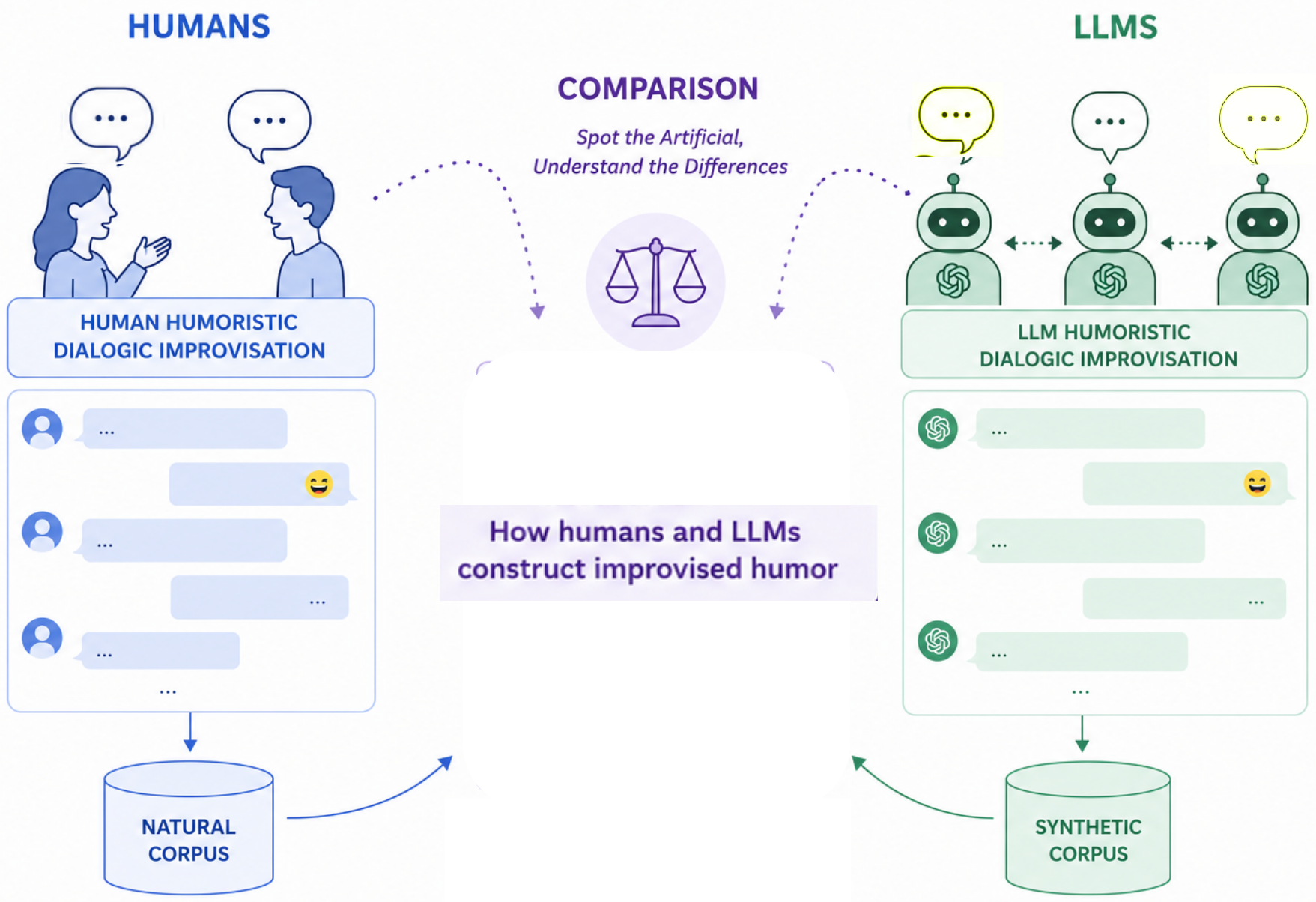}
    \caption{Overall framework for analyzing Dialogic Humor
Improvisation in Humans and LLMs}
    \label{fig:general}
\end{figure}

Our experimental setting consists of a comparative study of humorous improvisational dialogues produced by humans and by artificial agents. The human data are drawn from a television program featuring improvisational comedy, while the artificial setting is generated using large language models (LLMs). Specifically, we create a multi-agent environment in which each agent assumes the role of an actor engaged in humorous improvisation, generating responses based on the preceding interaction. See Figure \ref{fig:general}.

Furthermore, conducting a meaningful comparative analysis requires the incorporation of a linguistic perspective capable of capturing the subtle differences between natural and synthetic language. To this end, we integrate linguistic theory into our analytical framework, drawing on Systemic Functional Linguistics \cite{halliday1985introduction} as a foundation and complementing it with additional computational and qualitative tools. 

This interdisciplinary approach enables a more comprehensive examination of the linguistic mechanisms underlying both human, and LLM generated humorous interactions. This highlights the need of interdisicplinary methodologies  to address the challenges posed by contemporary AI systems and to achieve a deeper understanding of their  capabilities and limitations.

\section{Background}

In Natural Language Processing (NLP), a language model is a tool used to estimate the probability of sequences in a language and to generate text based on conditional probabilities. Current large language models (LLMs) are neural language models based on the Transformer architecture. Their scale, combined with vast amounts of training data and increasingly sophisticated architectures, enables them to capture patterns of language use and to construct abstract representations of meaning as well as certain forms of world knowledge. 

In this work, we use the term \textit{synthetic language} to refer to language produced by a large language model. We consider it synthetic in the sense that it follows the distributional patterns of natural language; however, the underlying mechanism that generates it is based on artificial neural networks rather than human cognitive processes. In contrast, we use the term \textit{spontaneous language} to refer to natural language produced by humans in a dialogic interaction setting.

\subsection{LLM-generated dialogues}

Research into large language models (LLMs) seeks to move beyond basic response generation toward emulating human-like dialogue through sophisticated persona creation and stylistic imitation, including phenomena like humor and sarcasm.

One first step towards creating human-like interaction is to integrate personas/characters into dialogue agents. For instance, some works propose to integrate core criteria when building a LLM-based character, e.g., Social Identity (demographics like age, gender, and occupation), Personal Identity (psychological traits and other values), and Personal Life Context (daily routines, preferences, and "likes/hates"). Research shows that including personal life context is the most effective element for making a character identifiable and realistic \cite{lee-etal-2025-spectrum}. Some works have used models like GPT-4o-mini to generate thousands of unique biographies based on datasets that contain over a billion personas. These personas can be assigned specific traits to ensure variety \cite{occhipinti-etal-2025-harry}. And finally, for domain-specific personas, agents are assigned roles such as "Physics-AI" or "Psychology-AI" to inject specific disciplinary perspectives into the conversation \cite{ueda-etal-2025-exploring}.

Phenomena such as humor have traditionally been addressed in NLP primarily as classification or detection tasks. However, the emergence of LLMs has expanded these capabilities, enabling the generation of humorous content rather than merely its identification. This shift opens new possibilities for studying how computational systems can reproduce complex communicative behaviors. In particular, prompting an LLM to imitate human dialogue traits, e.g., humor, requires a process of stylistic alignment: models are instructed to embody specific personas that include defined traits such as a "sarcastic sense of humor" or acting as a "consummate jokester" who uses one-liners. Successful imitation requires the model to adapt its language to reflect these character-specific quirks rather than providing generic helpful responses. Some works have identified that while zero-shot models can adapt to personas, they often exhibit "copy-paste" behavior, explicitly repeating biographical details in their dialogue \cite{russo2023countering,occhipinti-etal-2025-harry}. Fine-tuning on persona-based datasets is necessary to capture "deeper" traits—enabling the model to reflect a persona's values and experiences implicitly through its conversational tone and reactions \cite{occhipinti-etal-2025-harry}. Frameworks like DialogBench evaluate how well LLMs emulate human communication across tasks such as emotion detection, personality-grounded response generation, and understanding the interpersonal relationship between speakers \cite{ou-etal-2024-dialogbench}. 

Finally, an emerging area of investigation is how personas change depending on who they are speaking with. Investigating and implementing interlocutor adaptation and related tasks requires interdisciplinary approaches that combine linguistic theory and NLP expertise.

\subsection{Systemic Functional Linguistics (SFL)}
For the linguistic analysis, we adopt the framework of Systemic Functional Linguistics (SFL) \cite{halliday1985introduction}, focusing on the discourse-semantic stratum and, in particular, on the Appraisal Theory developed by \cite{martin2007language}. This framework provides a detailed account of the linguistic resources speakers use to express evaluation and to position themselves attitudinally, dialogically, and intertextually within discourse.

Appraisal Theory comprises three main subsystems: Attitude, Engagement, and Graduation. In this work we focus mainly on the \textbf{attitude} analysis; however we include a preliminary analysis of the other discoursive subsystems. 

\begin{itemize}
    \item \textbf{Attitude} is associated with utterances that convey a positive or negative evaluation. This category is divided into affect (the characterization of phenomena in relation to emotions), judgment (the evaluation of human behavior according to institutionalized social norms), and appreciation (the evaluation of objects, processes, and constructs in terms of aesthetic principles and social value).
    \item \textbf{Engagement} concerns the linguistic resources used to position the speaker’s or writer’s voice in relation to the propositions being communicated.  Utterances can be categorized in two types: monoglossic and heteroglossic. While monoglossic utterances present propositions without explicitly acknowledging alternative viewpoints or voices, heteroglossic utterances recognize the presence of other positions, perspectives, or possible interpretations.
    \item \textbf{Graduation} refers to a semantic domain of scaling that is concerned with how speakers intensify or diminish the force of their utterances and adjust the focus of their semantic categorizations, making them either sharper or more diffuse. Graduation is divided into two dimensions: force (which operates along a scale of intensity from low to high and is commonly realized through intensifiers, amplifiers, and emphatic adverbs) and focus (which indicates the degree to which a category is construed as central, prototypical, or marginal).  
\end{itemize}

We consider SFL a valuable theoretical bridge between linguistics and computational approaches to dialogue, providing an alternative methodological framework for the analysis of evaluative and interactional meanings in discourse.

To the best of our knowledge, little research has explored the integration of linguistic theory into the study of automatic humor generation. 

\section{Experimental setting}

\subsection{Spontaneous (natural) humorist dialogues}

To build the corpus of natural dialogues, we searched for publicly available examples of human theatrical improvisation centered on humor. The selected source was \textit{¿Quién dijo yo?}\footnote{https://www.youtube.com/watch?v=lsGrjoKo2E8}, a television improvisational comedy program in which actors perform spontaneous scenes based on prompts provided by the host and suggestions contributed by the audience. These audience contributions introduce unexpected elements that shape both the instructions and the resulting dialogues. The corpus is in Spanish.

For this study, we selected improvisation acts that could be reasonably replicated using conversational agents. Particular attention was given to scenarios involving creative prompts and discourse elements that contributed to humorous punchlines during the performance. The transcriptions were initially generated using automatic speech recognition tools\footnote{\url{https://huggingface.co/LuisG07/wav2vec2-large-xlsr-53-spanish}} and subsequently reviewed and corrected by a linguist to address any transcription errors.

\begin{table}
\centering
\caption{Spontaneous (natural) corpus. Each improvisation act represents a scenario introduced by the TV host, providing the prompt from which the actors develop their humorous improvised dialogues.}\label{tab:spontaneous}
\begin{tabular}{p{11.3cm} cp{1cm}}

\hline
\textbf{Improvisation act}    & \textbf{Turns} \\
\hline
\textit{Guess Who}\\ All situations take place inside a Mexico City taxi. The driver must take each passenger to their destination and, over the course of the conversation, guess which character the passenger is portraying & 82    \\
\hline
\textit{Four Scenarios}\\ A groom celebrates his bachelor party, which unexpectedly takes place across three locations: a dog grooming salon, a massage parlor, and a funeral home.                                              & 67    \\
\hline
\textit{Semantic Field}\\ The story takes place in an Argentine restaurant in Mexico, where a couple are celebrating their relationship. However, they unexpectedly encounter an ex-boyfriend                                  & 60    \\
\hline
\textit{Talk show}\\ The situation takes place on a talk show. The show's story centers on a man who, after suffering an injury while dancing, lost his career as a professional dancer.                                                     & 100   \\
\hline
\textbf{Total tokens:}         & 16931 \\
\hline
\end{tabular}
\end{table}

Due to the manual annotation/analysis required by our methodology, the corpus was limited to four improvisation acts, yielding a total of 309 natural dialogue turns. See Table \ref{tab:spontaneous}.

\subsection{Synthetic humorist dialogues}

In order to imitate the conversational acts observed in the spontaneous corpus, a straightforward approach would be to use an LLM and prompt it to follow the instructions of each improvisation act, generating a humorous dialogue among several actors. However, in this work, we argue that achieving greater fidelity to the conditions under which spontaneous conversations occur requires multiple agents to interact with one another. That is, each agent should generate its contributions based on the dialogue produced by the other agents. In this way, the LLM does not act as a scriptwriter generating the voices of multiple characters; instead, each agent faces the challenge of generating contextually appropriate responses based on the previous interventions of the other participants.

\begin{figure}[t]
    \centering
    \includegraphics[scale=0.17]{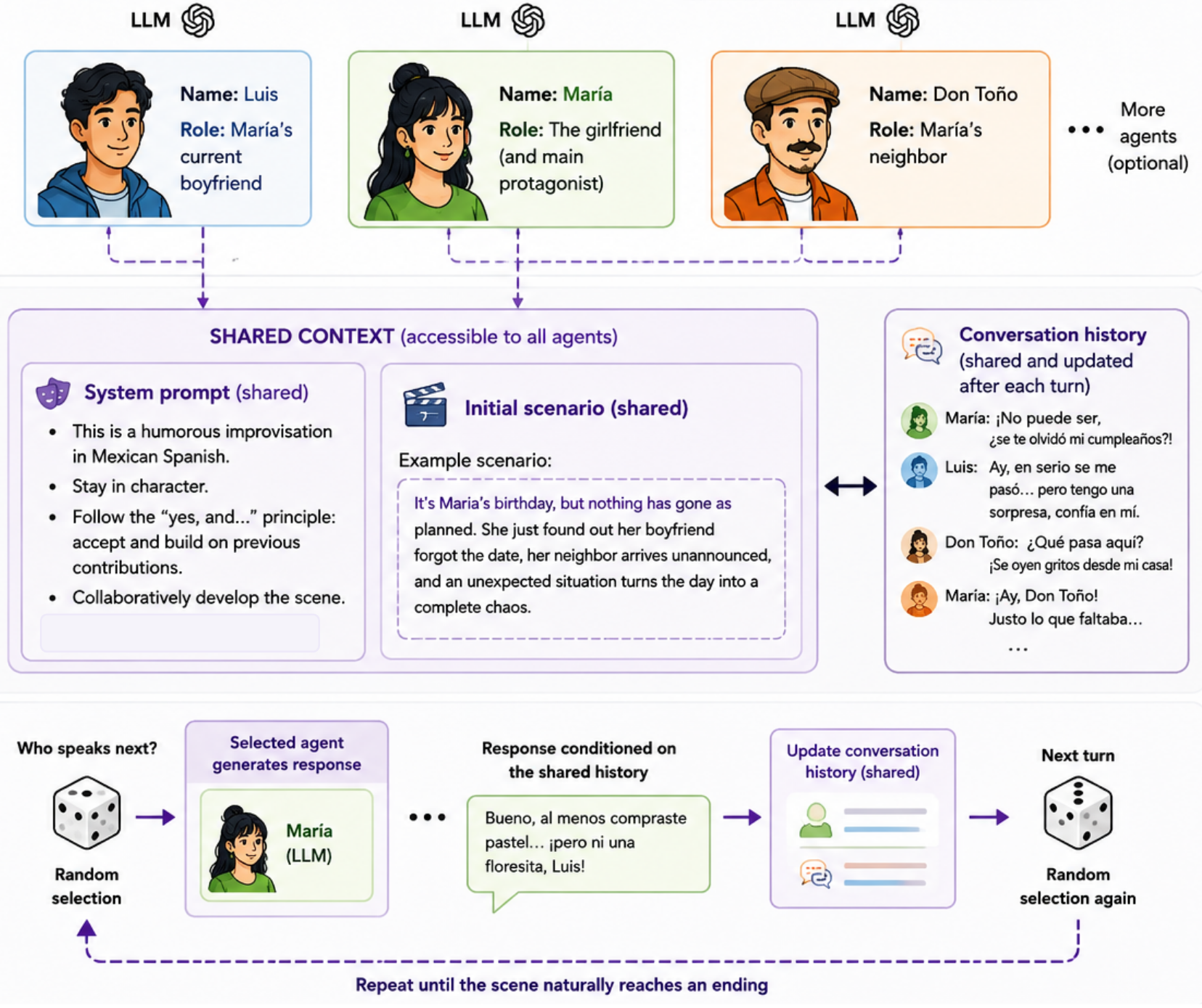}
    \caption{Proposed multi-agent setting for humorous dialogue improvisation}
    \label{fig:dialogues}
\end{figure}

Our experimental setup for generating synthetic dialogues with LLMs can be summarized as follows:

\begin{enumerate}
    \item We created several agents by initializing each LLM with a system prompt and assigning it a specific name and persona corresponding to a role in the improvisational sketch. For example, an agent could be instructed: "You are María's current boyfriend."
    \item Each agent operates independently; however, the conversation history is shared among all agents to ensure contextual consistency throughout the interaction. Part of this shared context consists of a system prompt specifying that the dialogue is a humorous improvisation performed in Mexican Spanish. The prompt instructs all agents to remain in character and to follow the theatrical improvisation principle of "yes, and...", whereby participants accept and build upon previous contributions to collaboratively develop the scene (see the Appendix \ref{app:prompt} for detailed prompts and model configurations).
    
    In addition, all agents receive the same description of the initial scenario, which serves as the starting point for the improvisation. This scenario provides the background and context from which the agents construct the dialogue, analogous to the information supplied by the host at the beginning of an improvisational television show performed by human actors.
    \item To imitate the spontaneity of natural dialogue, within the limits of LLMs, we do not predefine a fixed speaking order for the agents. Instead, we introduce a random component into the interaction process. At each conversational turn, any agent participating in the improvisational scene may be selected to continue the dialogue. Specifically, the next speaker is chosen at random, and the selected agent generates a response conditioned on the shared conversation history. This mechanism allows the dialogue to evolve dynamically and reduces the predictability that would arise from a predetermined turn-taking structure. See Figure \ref{fig:dialogues}.
    
\end{enumerate}
\textbf{Models}. For the instantiation of each agent, we used GPT-4o, a multimodal large language model developed by OpenAI. GPT-4o is a relatively recent model that offers strong performance across a wide range of natural language processing tasks while maintaining lower computational costs compared to larger alternatives. Its capabilities in dialogue generation, contextual reasoning, and role-playing make it particularly suitable for our improvisational setting. Furthermore, GPT-4o has been used in previous studies related to the generation of dialogues \cite{occhipinti-etal-2025-harry,lee-etal-2025-spectrum,qamar-etal-2025-llms}.

To define the scene, the roles of each agent, and the interaction process described above, we made use of system prompts, i.e., a specification that conditions the behavior of a language model. They function as a high-level instruction layer that defines the model's role, objectives, constraints, style, and behavioral guidelines before any interaction takes place. In all cases, the temperature of the model was set to 0.8 to encourage more creative and humorous responses.

Due to the stochastic nature of LLMs, we generated each humorous dialogue multiple times\footnote{10 times in most of the cases}. Each dialogue was generated with a reduced number of turns, in line with recent findings showing that persona alignment tends to decay through the conversational turns, with noticeable “instruction drift” occurring after 8 turns \cite{occhipinti-etal-2025-harry}.

It is important to note that, although LLMs can generate a large number of dialogues with relative ease, annotating them is considerably more challenging, especially when the analysis requires linguistic expertise that cannot be easily automated. For this reason, we selected one dialogue for each improvisation act for detailed manual annotation. This set constitutes our main synthetic corpus. Nevertheless, we retained all generated dialogues, allowing us to examine whether similar distributions and linguistic phenomena emerge across repeated generations despite the probabilistic nature of LLM outputs. This analysis can be conducted using statistical methods and automatic tools. Table \ref{tab:syncorpus} shows the information of the synthetic corpus.

\begin{table}
\centering
\caption{Synthetic corpus. Each improvisation act was generated using LLMs following the scenario and character roles defined for the scene. The table also reports the corpus size when dialogue generation is repeated multiple times.}\label{tab:syn}
\begin{tabular}{p{5.3cm} cp{1cm}}

\hline
\textbf{Improvisation act}    & \textbf{Turns} \\
\hline
\textit{Guess Who} & 36    \\
\textit{Four Scenarios}                                            & 12    \\
\textit{Semantic Field}& 12    \\
\textit{Talk show}  & 12   \\
\hline
\textbf{Total tokens:}         & 23037 \\
\hline
Total number of tokens across repeated dialogue generations. & 155178 \\
\hline
\end{tabular}
\label{tab:syncorpus}
\end{table}

\subsection{Comparative analyses}
Once the corpora had been constructed, comprising both spontaneous (natural) and synthetic humorous dialogues, we applied several methods to perform a comparative analysis. Our goal was to identify features that distinguish the two types of dialogue and, ultimately, to "spot the artificial" in order to better understand how humans and machines construct improvised humor.

\textbf{Analysis using automatic tools}. We analyze the length of dialogue turns in spontaneous human conversations and LLM-generated dialogues. Specifically, we measure turn length in terms of orthographic words and characters, and then compare the resulting distributions between natural and synthetic dialogues. Along with these measures, we also include the Moving-Average Type–Token Ratio (MATTR) as an indicator of lexical richness in the two corpora. MATTR is an enhanced version of the traditional Type–Token Ratio (TTR), designed to provide more reliable comparisons between corpora of different lengths. Instead of computing lexical diversity over the entire text, MATTR calculates the TTR within a moving window of fixed size and then averages the resulting values across all windows. 
\begin{equation}
\text{MATTR}(W) = \frac{1}{N-W+1}\sum_{i=1}^{N-W+1} \text{TTR}_i
\end{equation}
$$
\begin{aligned}
\textbf{Where:} & \\
N: & \quad \text{is the total number of tokens (words) in the text.} \\
W: & \quad \text{is the window size (the number of tokens in each sliding window).} \\
N - W + 1: & \quad \text{is the total number of sliding windows that can fit into the text.} \\
\text{TTR}_i: & \quad \text{is the Type-Token Ratio of the } i\text{-th window.} \\
\end{aligned}
$$
Furthermore, we use automatic taggers to analyze the corpora, focusing on part-of-speech (POS) tags, syntactic dependencies, and named entities (NERs). We compare the distributions of these linguistic features across the different corpora to determine whether there are significant differences in the use of grammatical categories, syntactic relations, and entity types between natural and synthetic dialogues. For performing POS tagging and dependency parsing we use \textit{Stanza} \cite{qi2020stanza}, NER is performed using \textit{GPT-5}.


\textbf{Linguistic analysis through human annotation}. Aside from the analyses described above, a qualitative analysis of the dialogues was conducted by a trained linguist using the Systemic Functional Linguistics (SFL) framework to examine discourse-level characteristics. Specifically, the annotation focused on identifying features belonging to the appraisal categories of attitude. For example, within the attitude system, instances of affect, judgment, and appreciation can be identified in the dialogues. This annotation was performed manually for both the spontaneous (natural) corpus and the synthetic corpus. A preliminary qualitative analysis of the engagement and graduation discoursive systems is also presented.

\section{Results and discussion}

\begin{figure}
    \includegraphics[width=1\linewidth]{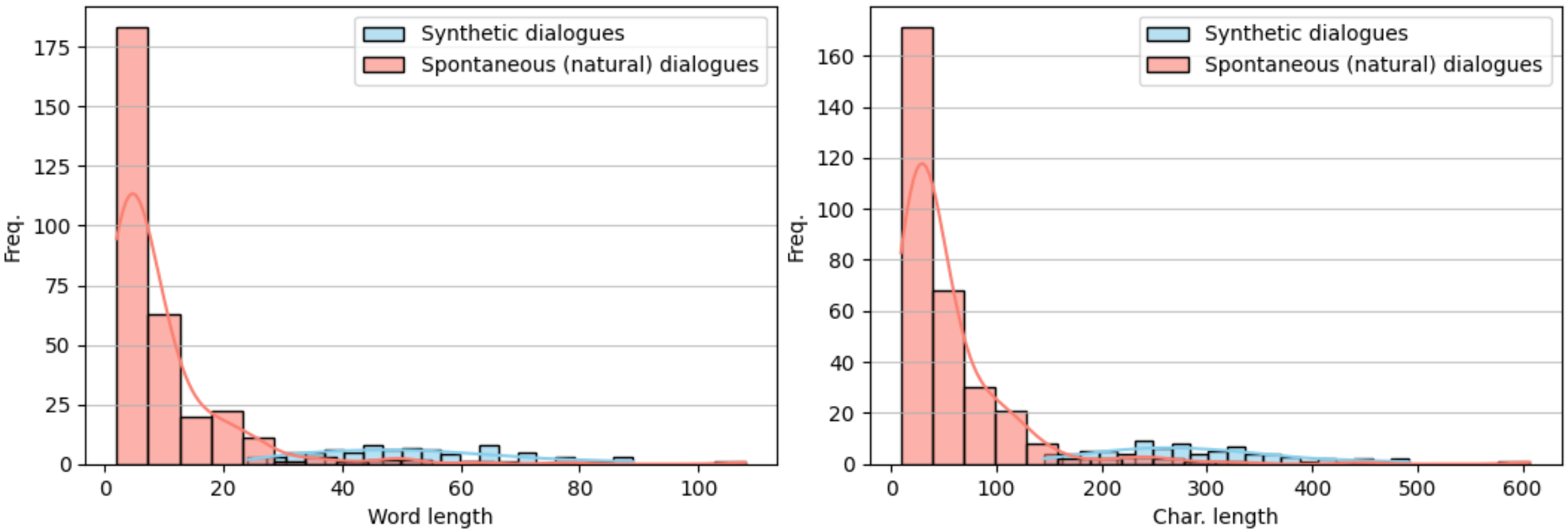}
    \caption{Histograms of conversational turn lengths in words and characters.}
    \label{fig:length}
\end{figure}

\textbf{Turn Length.} We compare the length of turns in the synthetic and natural corpora. One clear observation is that turns in the synthetic dialogues tend to be considerably longer than those in their spontaneous counterparts. This difference may be related to the fact that the natural corpus reflects the dynamics of spoken interaction, including interruptions, short interjections, and rapid speaker exchanges. In contrast, turns in the synthetic dialogues tend to be more narrative and self-contained, resulting in fewer rapid shifts between speakers and longer conversational contributions. As shown in Figure~\ref{fig:length}, the length of most turns in the natural corpus ranges from approximately 5 to 20 words, whereas turns in the synthetic dialogues are primarily concentrated between 40 and 80 words. In fact, the average turn length is 9.42 words in the natural corpus and 51 words in the synthetic corpus, revealing a substantial difference in the construction of humorous dialogue across the two settings. A similar pattern can be observed when turn length is measured in terms of characters.

\textbf{Type-token ratio}
Consistent with the previous observations, the MATTR analysis, computed using different window sizes, showed that the synthetic dialogues consistently exhibit higher values than the natural corpus. This suggests greater lexical diversity in the synthetic dialogues. We interpret this finding as another indication that synthetic dialogues struggle to capture the spontaneity of natural dialogic interactions. Previous research has shown that spoken conversation tends to exhibit lower lexical diversity than written language due to real-time cognitive and interactional constraints \cite{891328b8168b4c72953b72c0e7b1d7ee}. Speakers often rely on repetition, discourse routines, shared context, and the reuse of lexical material introduced by other interlocutors, all of which contribute to reduced lexical diversity \cite{koizumi2012effects,clavel2021analysing}. In this sense, the LLM-based agents do not appear to fully mimic the type of turns observed in spontaneous conversations, which are typically shorter, more dynamic, and characterized by lower lexical diversity.

\begin{table}[]
\centering
\caption{Moving average type token ratio (MATTR) across different window sizes for the two corpora}\label{tab:ttr}
\begin{tabular}{|l|cccc|}
\hline
                      & \multicolumn{4}{c|}{\textbf{Window size}}                                                \\ \hline
\textbf{MATTR}        & \multicolumn{1}{c|}{\textbf{25}}   & \multicolumn{1}{c|}{\textbf{50}}   & \multicolumn{1}{c|}{\textbf{100}}  & \textbf{150}  \\ \hline
Spontaneous (natural) & \multicolumn{1}{c|}{0.80} & \multicolumn{1}{c|}{0.71} & \multicolumn{1}{c|}{0.62} & 0.56 \\ \hline
Synthetic             & \multicolumn{1}{c|}{0.91} & \multicolumn{1}{c|}{0.82} & \multicolumn{1}{c|}{0.70} & 0.62 \\ \hline
\end{tabular}
\end{table}

It is worth noting that these trends remain consistent across the multiple synthetic corpora generated through repeated runs (see Appendix \ref{app:plots}), suggesting that the observations are not merely an artifact of the particular synthetic corpus selected for analysis.

\begin{figure}
    \includegraphics[width=1\linewidth]{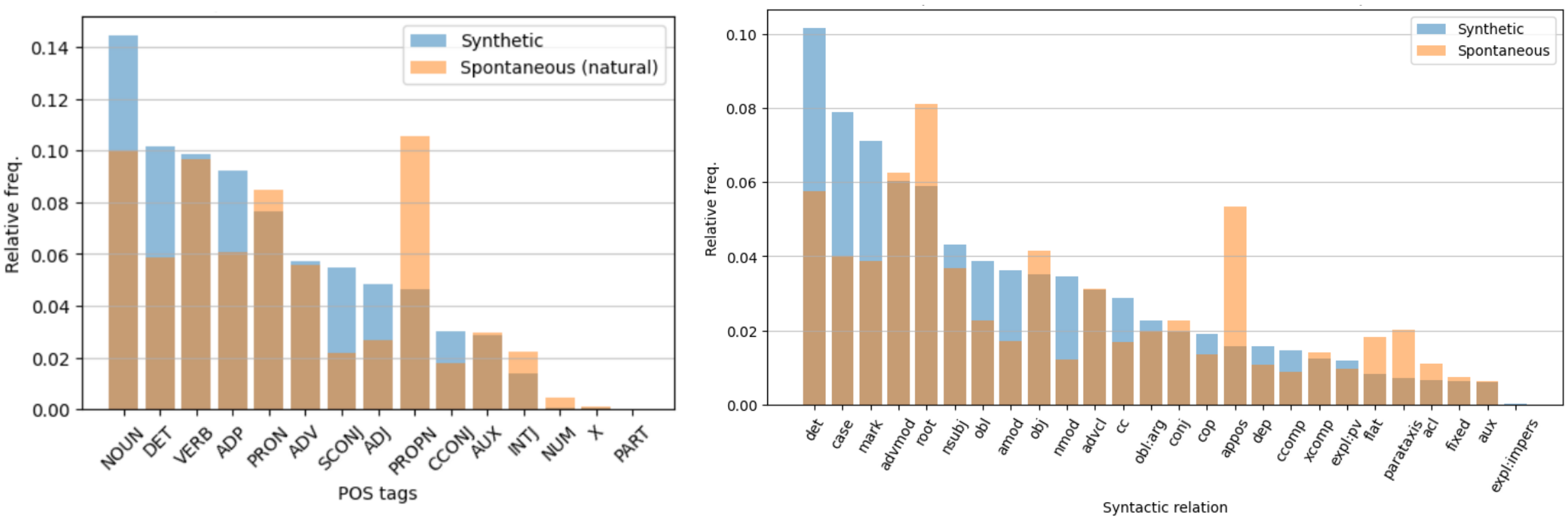}
    \caption{Distribution of POS tags and syntactic dependencies across corpora}
    \label{fig:pos}
\end{figure}

\textbf{POS tags, syntactic dependencies and NERs}. In Figure \ref{fig:pos}, we can observe the relative frequencies of POS tags and syntactic relations across the two corpora. While several categories exhibit similar distributions, some show pronounced differences. As a general observation, the synthetic dialogues appear to be more syntactically organized. Although the agents are instructed to improvise, they do not fully reproduce the characteristics of spontaneous conversational speech. As a result, the language generated by LLMs lacks some of the disfluencies, interruptions, and irregular constructions that are common in human improvisation.

For example, the synthetic corpus shows a higher frequency of adjectives and adjectival modifiers \textit{(amod)} compared to the spontaneous dialogues. This pattern suggests not only a greater use of descriptive language, but also a tendency toward more elaborated and syntactically structured utterances.

Another example is the use of determiners \textit{(det)}, which are more frequent in the synthetic corpus. This suggests that LLMs tend to generate more complete and elaborated noun phrases, whereas spontaneous speech often contains shorter and less explicit nominal constructions. In addition, the synthetic dialogues exhibit a higher frequency of conjunctions, reflecting a more structured discourse organization. This pattern may indicate a greater use of subordinate and coordinated clauses, as also suggested by the increased frequency of coordinating conjunctions \textit{(cc)}.

By contrast, a characteristic feature of spontaneous speech is the more frequent use of parataxis, that is, clauses that can stand independently without relying on other clauses for their interpretation. The natural corpus shows a higher frequency of this relation, consistent with the shorter, less structured, and more fragmented utterances that typically occur in spontaneous conversational interaction.

\begin{figure}
   \centering
    \includegraphics[scale=0.12]{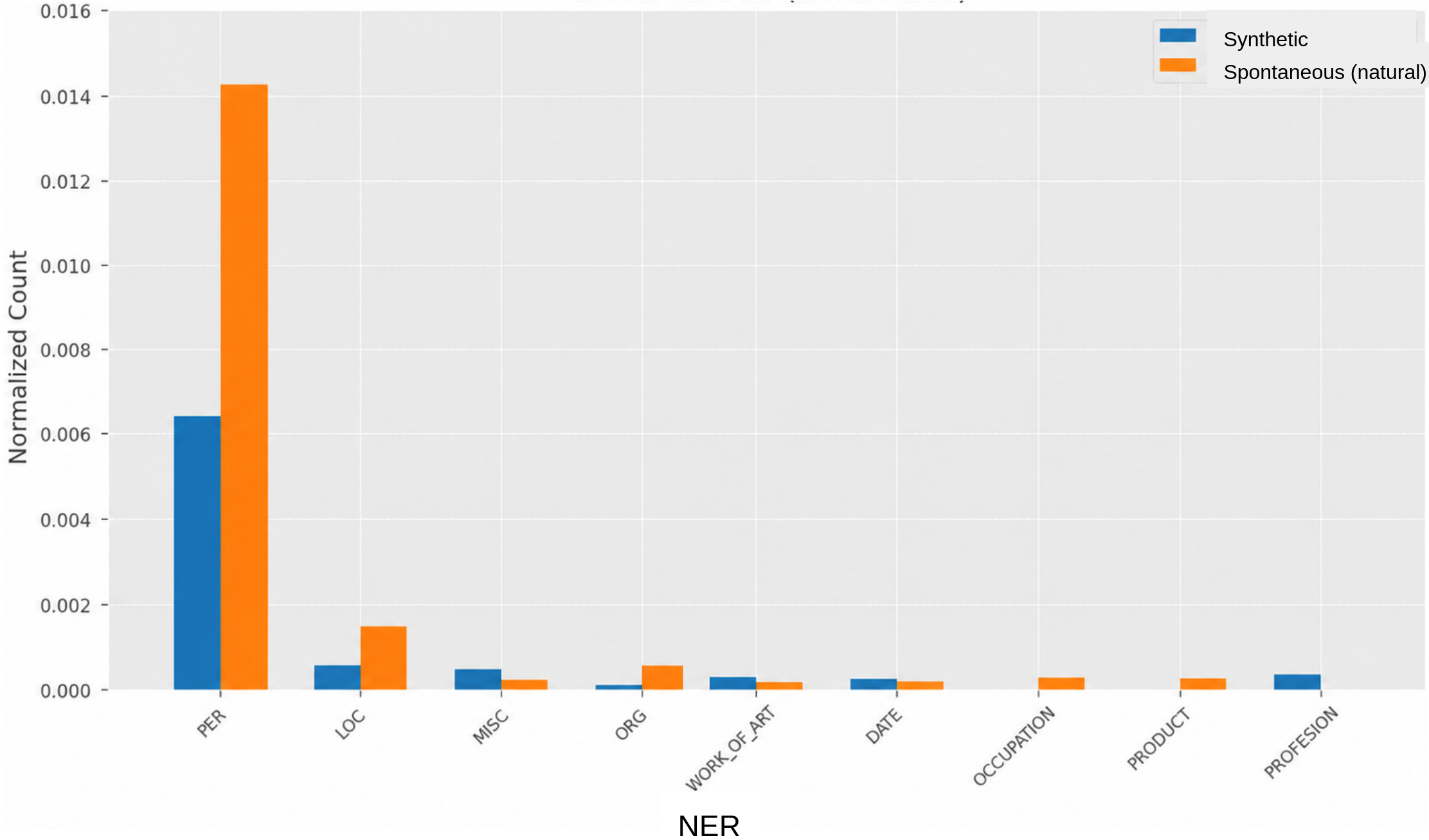}
    \caption{NER distributions across corpora.}
    \label{fig:ner}
\end{figure}

A POS tag that shows a marked contrast between the two types of corpora is the proper nouns (\textit{propn}). A deeper analysis using named entities recognition revealed that both synthetic and spontaneous humorous dialogues exhibit a similar distribution of named entities, with person and location entities being the most common. In contrast, other domains, such as news text, typically contain a wider variety of named entities. However, despite their similar overall distribution of named entities, the spontaneous dialogues contain a higher relative frequency of person and location entities. This suggests that, although synthetic dialogues are longer and contain more tokens, they rely less on proper names and location references than their spontaneous counterparts. See Figure \ref{fig:ner}.

The higher frequency of named entities/proper nouns in spontaneous dialogues may reflect the need to continuously identify and emphasize participants, locations, and other referents during real-time interaction. Speakers often rely on references to people, places, and institutions to construct and maintain a shared conversational context. In contrast, synthetic dialogues tend to present information in a more organized manner, reducing the need for explicit references to interlocutors and contextual elements. They may achieve this by using other resources like pronouns and clitics. In sum, synthetic language may appear more carefully structured, while the spoken interaction that is reflected in spontaneous corpus remains a highly complex communicative process that relies heavily on contextual and interactional cues.

\textbf{Discourse-level characteristics.} Using the appraisal theory of SFL, we performed a comparative analysis across the different subsystems. This required manual annotation and assessment by an expert linguist. Here we syntheize some of the main observations:

The distribution of attitude types reveals a marked difference between spontaneous and synthetic dialogues. In spontaneous human improvisations, affect is the most frequent type of attitudinal resource, followed by appreciation and judgment. This distribution suggests that human participants rely heavily on the expression of emotions, feelings, and affective responses when constructing humorous interactions. In contrast, synthetic dialogues show a substantial shift toward appreciation, relying more strongly on evaluations of entities, situations, and experiences than on the direct expression of emotion. Regarding judgment, there is a relatively similar proportion across both corpora, which indicates that evaluations of social behavior and personal characteristics constitute a comparatively stable component of the attitude system. See Table \ref{tab:proportion}.

\begin{table}[]
\center
\begin{tabular}{|l|l|l|}
\hline
                      & \textbf{Natural} & \textbf{Synthetic} \\ \hline
\textbf{Affect}       & 55.6\%           & 32.8\%             \\
\textbf{Appreciation} & 24.4\%           & 51.1\%             \\
\textbf{Judgement}    & 20\%             & 16.1\%   \\\hline    

\end{tabular}
\caption{Total percentage of elements found in the attitude subsystem}
\label{tab:proportion}
\end{table}

\begin{figure}
   \centering
    \includegraphics[scale=0.11]{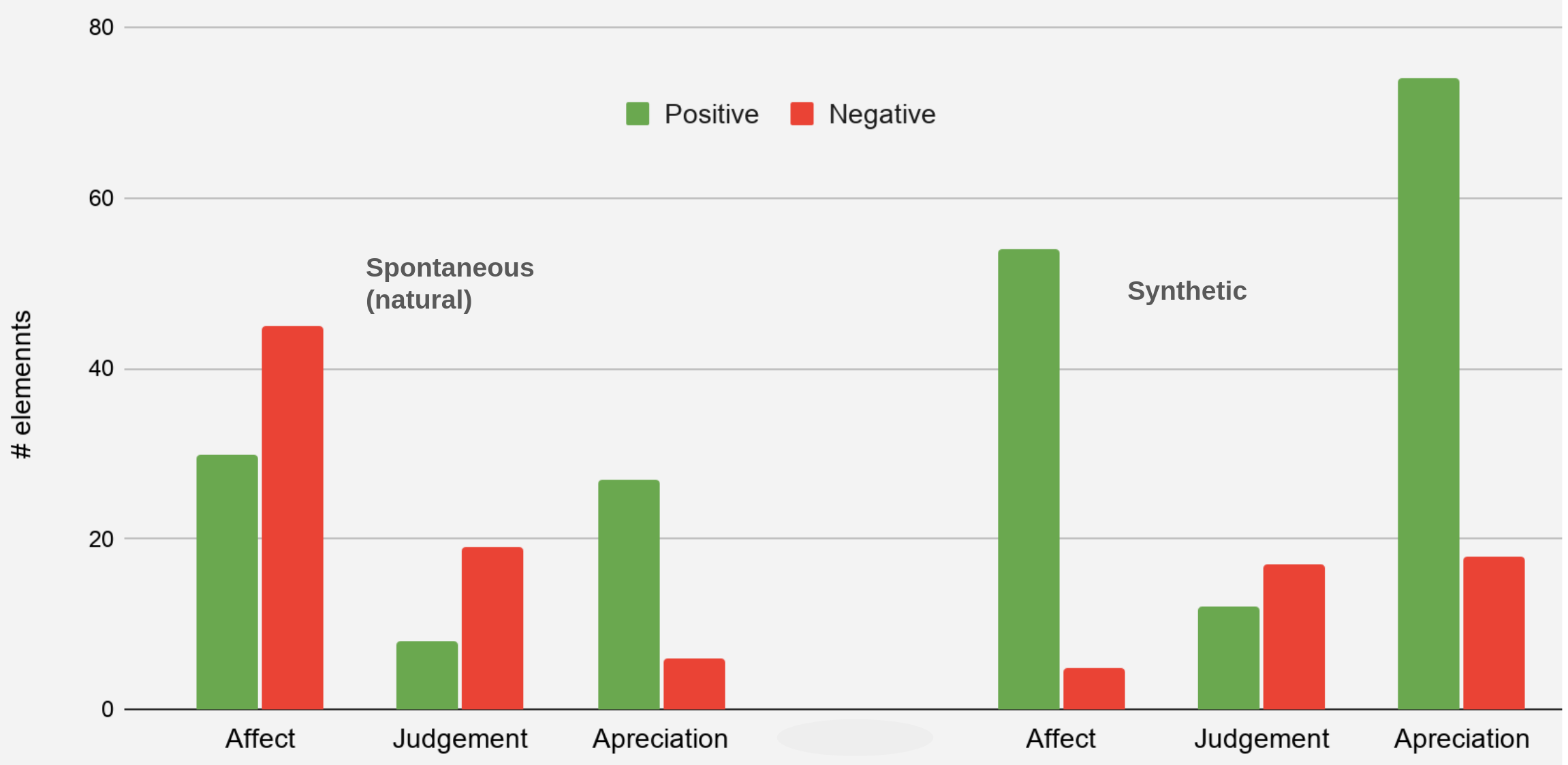}
    \caption{Analysis of the attitudes system based on the appraisal theory}
    \label{fig:attitude}
\end{figure}

Furthermore, human improvisations tend to exhibit a more balanced distribution of positive and negative evaluations, expressed through affect, judgment, and appreciation. In contrast, LLM-generated dialogues show a stronger tendency toward positive evaluations, often expressing supportive, compassionate, and agreeable attitudes. Additionally, human speakers in humorous improvisation appear less concerned with maintaining a socially desirable image, freely producing both favorable and unfavorable judgments. This tendency might be related to the alignment mechanisms used in LLMs, which encourage polite and socially acceptable interactions. Their bias toward positive responses, and in some cases sycophantic behavior, has been documented in several studies \cite{tian2024large,sun2026friendly}. See Figure \ref{fig:attitude} for detailed results.

Another interesting difference is that spontaneous dialogues rarely contain complex rhetorical constructions such as metaphors, similes, or elaborate comparisons, likely due to the real-time constraints of improvisation. Synthetic dialogues, however, make more frequent use of these devices, reflecting their greater capacity for planning and linguistic elaboration.

In a more qualitative fashion, our preliminary analysis suggests that, in terms of \textbf{engagement}, spontaneous improvisation is inherently heteroglossic, with participants continuously expanding on and challenging each other's ideas. Overall, spontaneous dialogues exhibit a more open form of interpersonal negotiation, whereas synthetic dialogues tend to construct more contrastive and less collaborative positions. This does not mean that the dialogic exchange is closed, but rather that the synthetic interactions struggle to consider other voices during the interaction. 

Similarly, in terms of \textbf{Graduation}, we observed that spontaneous language exhibits a greater use of intensification because, in the fluency of speech, there is no planning, and speakers resort to constant repetition of elements to generate an idea. In contrast, synthetic language is more structured and does not depend on immediacy; therefore, its use is balanced between intensification and quantification. A more fine-grained analysis of these subsystems is left for future work.

Finally, the prosodic elements of spontaneous conversation are particularly difficult for LLMs to imitate. Although textual devices such as emojis, exclamation marks, or unconventional orthography can sometimes be used to create a more dynamic conversational space, they are unable to fully reproduce this distinctly human aspect of interaction.

It is also important to note that, overall, we observed that although LLMs show creativity in developing scenes with coherent openings and closures, they tend to behave more like scriptwriters than conversational participants. During generation, it was often necessary to remind the agents through system prompts to remain in character and avoid producing the dialogue of other participants. This suggests that current\textbf{ LLMs are better at writing dialogues than at engaging in spontaneous conversational interaction}.

\section{Conclusions}
This work proposed a methodology for examining the differences between natural and synthetic language in a particularly challenging setting: humorous improvisational dialogue. We presented a framework for recreating this environment using LLM-based agents. Our comparative approach combines automated tools with manual annotation grounded in linguistic theory, enabling the analysis of both surface-level patterns and discourse-level characteristics.

Our findings suggest that, although LLMs can act as creative writers and successfully develop scenes with coherent openings and closings, they often fail to reproduce key features of human spoken improvisation. In particular, human dialogues tend to contain shorter, more dynamic turns and exhibit less rigid syntactic organization. By contrast, LLM-generated language remains highly structured and grammatically well-formed, even when agents are instructed to improvise. The analysis also revealed that LLMs struggle to produce inappropriate, confrontational, or negative evaluations, features that are often present and even necessary in human humorous improvisation. Instead, synthetic dialogues consistently display a tendency towards positive evaluations.

To the best of our knowledge, little research has explored the integration of linguistic theory into the study of automatic humor generation. This gap highlights the value of interdisciplinary approaches that combine insights from linguistics and NLP to better understand how humor is constructed in both human and machine-generated interactions. Although our results show that current LLMs struggle to reproduce several characteristics of human dialogic improvisation, this does not imply that they cannot improve substantially through instruction, adaptation, or fine-tuning. Rather, our findings emphasize that rigorous linguistic analysis is a necessary first step for identifying the aspects that require improvement, demonstrating the important role that linguistic expertise can play in the development and evaluation of future conversational AI systems.

\begin{credits}
\subsubsection{\ackname} 
We thank UNAM’s PAPIIT project TA100725 and SECIHTI’s master’s scholarship (2022537) for their financial support.

\end{credits}


\bibliographystyle{splncs04}
\bibliography{mybibliography}

@misc{attardo2008semantics,
  title={Semantics and pragmatics of humor. Language and Linguistics Compass, 2 (6), 1203--1215},
  author={Attardo, S},
  year={2008}
}

@article{891328b8168b4c72953b72c0e7b1d7ee,
title = "Toward a crosslinguistic comparison of lexical quanta in speech and writing",
author = "Sven Str{"o}mqvist and Victoria Johansson and Sarah Kriz and Hrafnhildur Ragnarsd{'o}ttir and Ravid Aisenman and Dorit Ravid",
note = "The information about affiliations in this record was updated in December 2015. The record was previously connected to the following departments: Linguistics and Phonetics (015010003)",
year = "2002",
language = "English",
volume = "5",
pages = "45--67",
journal = "Written Language and Literacy",

}

@inproceedings{tian2024large,
  title={Are large language models capable of generating human-level narratives?},
  author={Tian, Yufei and Huang, Tenghao and Liu, Miri and Jiang, Derek and Spangher, Alexander and Chen, Muhao and May, Jonathan and Peng, Nanyun},
  booktitle={Proceedings of the 2024 Conference on Empirical Methods in Natural Language Processing},
  pages={17659--17681},
  year={2024}
}

@inproceedings{sun2026friendly,
  title={Be friendly, not friends: How llm sycophancy shapes user trust},
  author={Sun, Yuan and Wang, Ting},
  booktitle={Proceedings of the 2026 CHI Conference on Human Factors in Computing Systems},
  pages={1--15},
  year={2026}
}

@inproceedings{russo2023countering,
  title={Countering misinformation via emotional response generation},
  author={Russo, Daniel and Kaszefski-Yaschuk, Shane and Staiano, Jacopo and Guerini, Marco},
  booktitle={Proceedings of the 2023 Conference on Empirical Methods in Natural Language Processing},
  pages={11476--11492},
  year={2023}
}

@book{halliday1985introduction,
  title={An Introduction to Functional Grammar},
  author={Halliday, M.A.K.},
  isbn={9780713163650},
  lccn={gb84033956},
  year={1985},
  publisher={E. Arnold}
}

@book{martin2007language,
  title={The Language of Evaluation: Appraisal in English},
  author={Martin, J. and White, P.R.R.},
  isbn={9780230511910},
  url={https://books.google.cl/books?id=I-J8DAAAQBAJ},
  year={2007},
  publisher={Palgrave Macmillan UK}
}

@inproceedings{qi2020stanza,
    title={Stanza: A {Python} Natural Language Processing Toolkit for Many Human Languages},
    author={Qi, Peng and Zhang, Yuhao and Zhang, Yuhui and Bolton, Jason and Manning, Christopher D.},
    booktitle = "Proceedings of the 58th Annual Meeting of the Association for Computational Linguistics: System Demonstrations",
    year={2020}
}

@article{clavel2021analysing,
  title={Analysing lexical density, diversity, and sophistication in written and spoken telecollaborative exchanges},
  author={Clavel-Arroitia, Bego{\~n}a and Pennock-Speck, Barry},
  journal={Computer-Assisted Language Learning Electronic Journal},
  volume={22},
  number={3},
  pages={230--250},
  year={2021}
}

@article{koizumi2012effects,
  title={Effects of text length on lexical diversity measures: Using short texts with less than 200 tokens},
  author={Koizumi, Rie and In'nami, Yo},
  journal={System},
  volume={40},
  number={4},
  pages={554--564},
  year={2012},
  publisher={Elsevier}
}

@inproceedings{qamar-etal-2025-llms,
    title = "Do {LLM}s Understand Dialogues? A Case Study on Dialogue Acts",
    author = "Qamar, Ayesha  and
      Tong, Jonathan  and
      Huang, Ruihong",
    editor = "Che, Wanxiang  and
      Nabende, Joyce  and
      Shutova, Ekaterina  and
      Pilehvar, Mohammad Taher",
    booktitle = "Proceedings of the 63rd Annual Meeting of the Association for Computational Linguistics (Volume 1: Long Papers)",
    month = jul,
    year = "2025",
    address = "Vienna, Austria",
    publisher = "Association for Computational Linguistics",
    url = "https://aclanthology.org/2025.acl-long.1271/",
    doi = "10.18653/v1/2025.acl-long.1271",
    pages = "26219--26237",
    ISBN = "979-8-89176-251-0"
}

@inproceedings{ou-etal-2024-dialogbench,
    title = "{D}ialog{B}ench: Evaluating {LLM}s as Human-like Dialogue Systems",
    author = "Ou, Jiao  and
      Lu, Junda  and
      Liu, Che  and
      Tang, Yihong  and
      Zhang, Fuzheng  and
      Zhang, Di  and
      Gai, Kun",
    editor = "Duh, Kevin  and
      Gomez, Helena  and
      Bethard, Steven",
    booktitle = "Proceedings of the 2024 Conference of the North American Chapter of the Association for Computational Linguistics: Human Language Technologies (Volume 1: Long Papers)",
    month = jun,
    year = "2024",
    address = "Mexico City, Mexico",
    publisher = "Association for Computational Linguistics",
    url = "https://aclanthology.org/2024.naacl-long.341/",
    doi = "10.18653/v1/2024.naacl-long.341",
    pages = "6137--6170"
}

@inproceedings{occhipinti-etal-2025-harry,
    title = "When Harry Meets Superman: The Role of The Interlocutor in Persona-Based Dialogue Generation",
    author = "Occhipinti, Daniela  and
      Guerini, Marco  and
      Nissim, Malvina",
    editor = "Che, Wanxiang  and
      Nabende, Joyce  and
      Shutova, Ekaterina  and
      Pilehvar, Mohammad Taher",
    booktitle = "Proceedings of the 63rd Annual Meeting of the Association for Computational Linguistics (Volume 1: Long Papers)",
    month = jul,
    year = "2025",
    address = "Vienna, Austria",
    publisher = "Association for Computational Linguistics",
    url = "https://aclanthology.org/2025.acl-long.879/",
    doi = "10.18653/v1/2025.acl-long.879",
    pages = "17964--17985",
    ISBN = "979-8-89176-251-0"
}

@inproceedings{ueda-etal-2025-exploring,
    title = "Exploring the Design of Multi-Agent {LLM} Dialogues for Research Ideation",
    author = "Ueda, Keisuke  and
      Hirota, Wataru  and
      Takahashi, Kosuke  and
      Omi, Takahiro  and
      Arima, Kosuke  and
      Ishigaki, Tatsuya",
    editor = "B{\'e}chet, Fr{\'e}d{\'e}ric  and
      Lef{\`e}vre, Fabrice  and
      Asher, Nicholas  and
      Kim, Seokhwan  and
      Merlin, Teva",
    booktitle = "Proceedings of the 26th Annual Meeting of the Special Interest Group on Discourse and Dialogue",
    month = aug,
    year = "2025",
    address = "Avignon, France",
    publisher = "Association for Computational Linguistics",
    url = "https://aclanthology.org/2025.sigdial-1.26/",
    pages = "322--337"
}

@inproceedings{lee-etal-2025-spectrum,
    title = "{SP}e{C}trum: A Grounded Framework for Multidimensional Identity Representation in {LLM}-Based Agent",
    author = "Lee, Keyeun  and
      Kim, Seo Hyeong  and
      Lee, Seolhee  and
      Eun, Jinsu  and
      Ko, Yena  and
      Jeon, Hayeon  and
      Kim, Esther Hehsun  and
      Cho, Seonghye  and
      Yang, Soeun  and
      Kim, Eun-mee  and
      Lim, Hajin",
    editor = "Chiruzzo, Luis  and
      Ritter, Alan  and
      Wang, Lu",
    booktitle = "Proceedings of the 2025 Conference of the Nations of the Americas Chapter of the Association for Computational Linguistics: Human Language Technologies (Volume 1: Long Papers)",
    month = apr,
    year = "2025",
    address = "Albuquerque, New Mexico",
    publisher = "Association for Computational Linguistics",
    url = "https://aclanthology.org/2025.naacl-long.356/",
    doi = "10.18653/v1/2025.naacl-long.356",
    pages = "6971--6991",
    ISBN = "979-8-89176-189-6"
}

\newpage
\appendix

\section{System prompts} \label{app:prompt}
\begin{verbatim}
shared_history = [
        {"role": "system", "content": f""" Esta es una improvisación humorística en 
        
        español mexicano. Todos los agentes deben improvisar una conversación y
        
        mantenerse en personaje, deben seguir la regla 'sí y además...' (yes-and) de
        
        de la improvisación  teatral. Deben tratar de ser graciosos. REGLA ESTRICTA: 
        
        cada agente debe generar su línea de diálogo, no actuar como narrador ni 
        
        generar diálogo de otros  personajes.  El escenario es: {scenario}"""}
    ]
\end{verbatim}

\begin{verbatim}
 # Define the agents with their names and personas.
    agents = [
        Agent(
            name="Gladys",
            persona="Joven novia de Rodolfo, su ex-novio d ela secundaria fue José"
        ),
        Agent(
            name="Rodolfo",
            persona="Novio actual de Gladys"
        ),
        Agent(
            name="José",
            persona="Joven que fue novio de Gladys en la secundaria y le gusta 
            
            recordar viejos tiempos"
        )
    ]

    # Define the scenario for the improv sketch.
    scenario = "Un restaurante argentino donde una pareja de novios, Gladys y 
    
    Rodolfo, van a celebrar el inicio de su noviazgo. Sin embargo, se encuentran a 
    
    José, el ex-novio de Gladys"

    \end{verbatim}

\section{Analysis using all synthetic corpora} \label{app:plots}

\begin{figure}
    \includegraphics[width=1\linewidth]{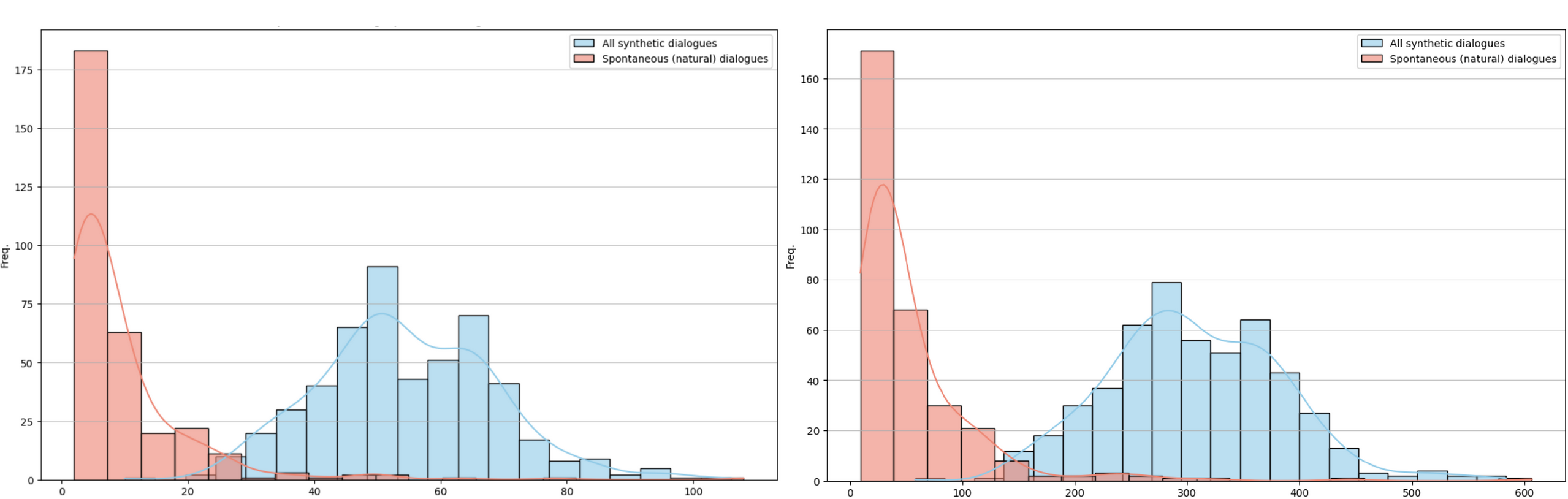}
    \caption{Histograms of conversational turn lengths in words and characters.}
    \label{fig:length2}
\end{figure}
\begin{figure}
    \includegraphics[scale=0.13]{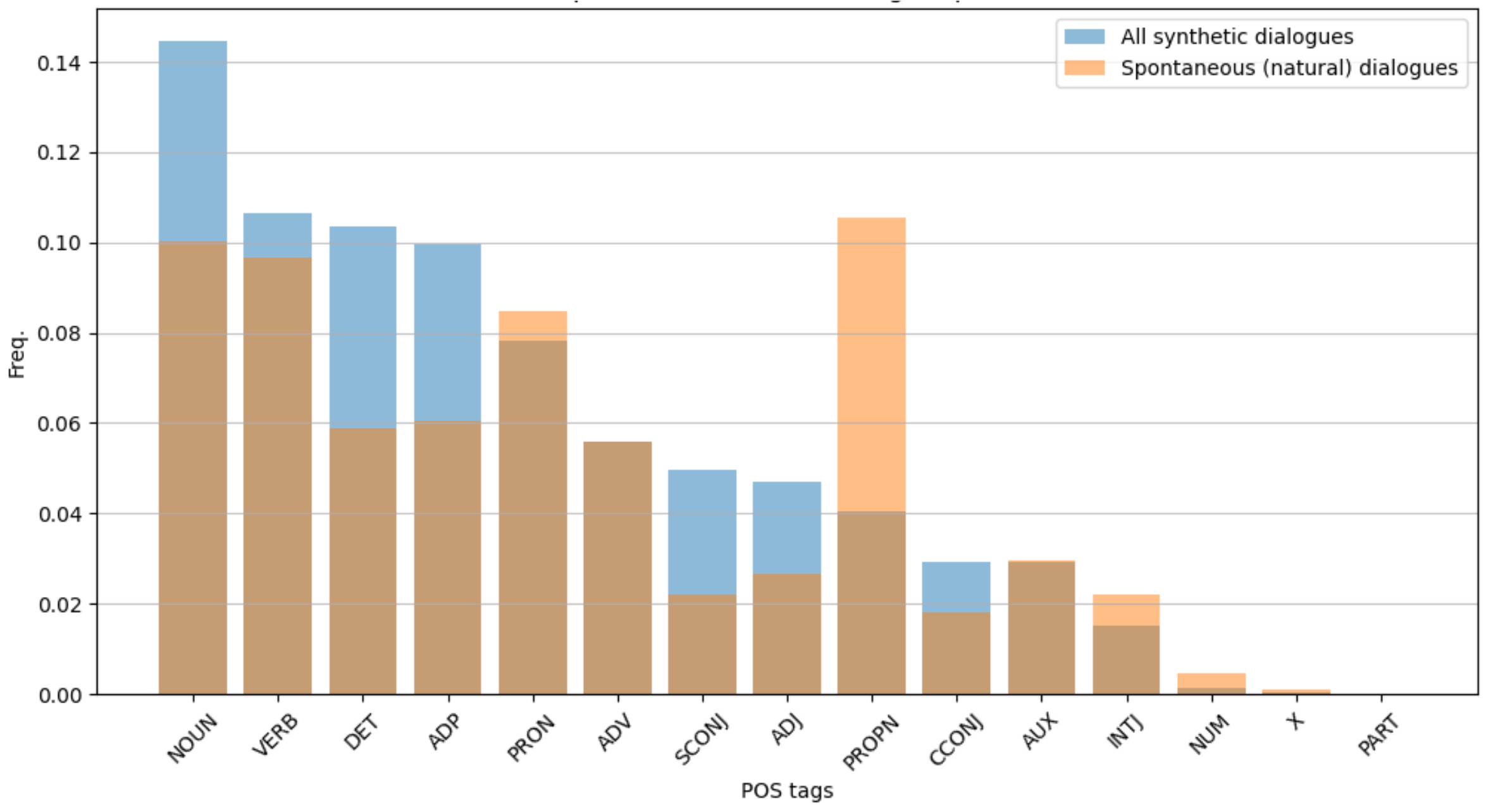}
    \caption{Distribution of POS tags across corpora}
    \label{fig:pos2}
\end{figure}

\begin{figure}
    \includegraphics[scale=0.14]{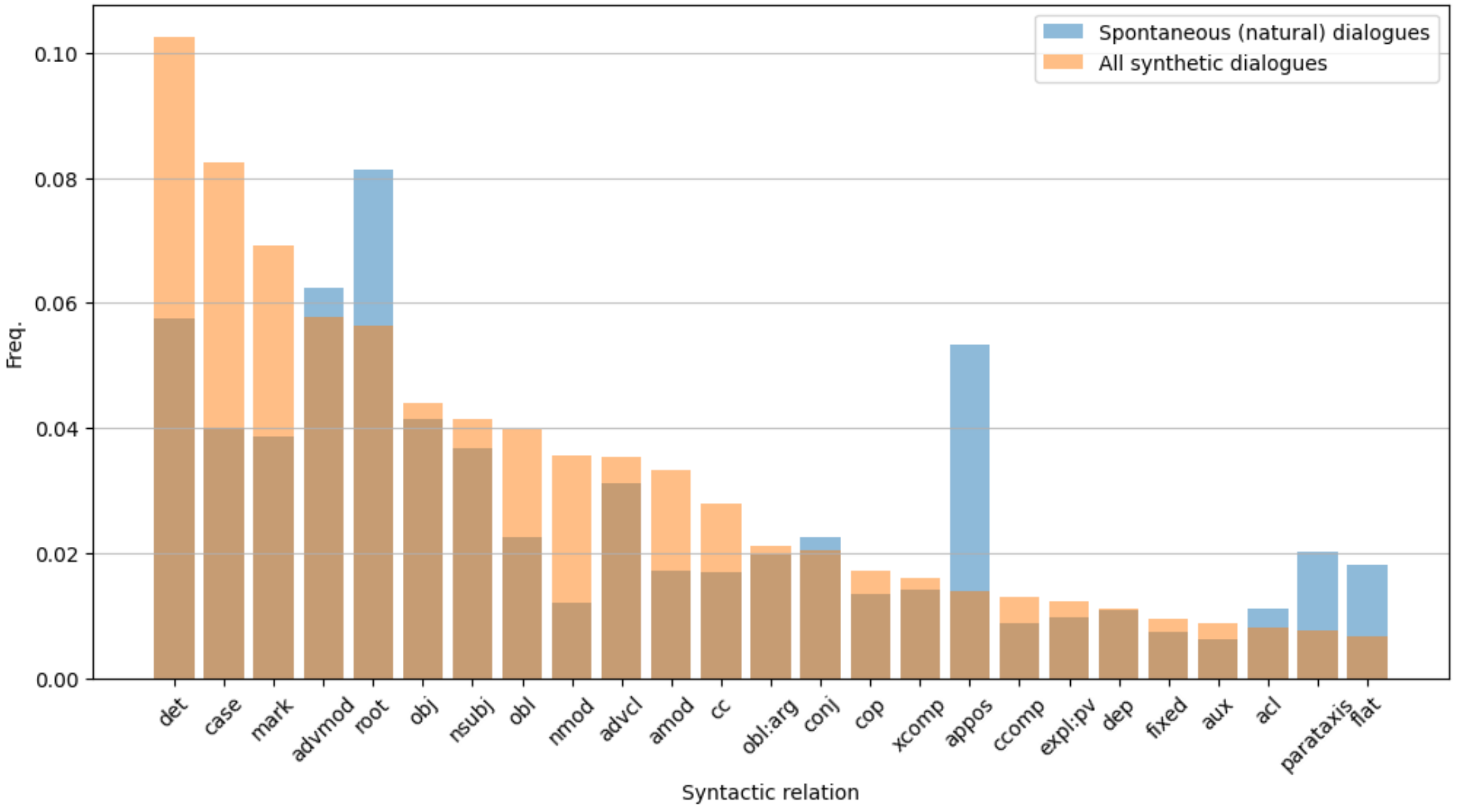}
    \caption{Distribution of syntactic dependencies across corpora}
    \label{fig:pos3}
\end{figure}

\end{document}